\documentclass[conference]{IEEEtran}

\usepackage{epstopdf}%
\usepackage[caption=false]{subfig}%

\usepackage[usenames,dvipsnames]{color}

\usepackage[pdftex]{graphicx}
\usepackage{amsthm}
\usepackage{url}
\usepackage{amsmath}
\usepackage{cite}
\usepackage{bmpsize}

\theoremstyle{plain}%

\theoremstyle{definition}

\theoremstyle{remark}

\usepackage{enumitem}

\begin{document}

\title{Synthetic Lung Nodule 3D Image Generation Using Autoencoders}

\author{\IEEEauthorblockN{Steve Kommrusch}
\IEEEauthorblockA{\textit{Dept. of Computer Science} \\
\textit{Colorado State University}\\
Fort Collins, Colorado, USA \\
steve.kommrusch@gmail.com}
\and
\IEEEauthorblockN{Louis-No{\"e}l Pouchet}
\IEEEauthorblockA{\textit{Dept. of Computer Science} \\
\textit{Colorado State University}\\
Fort Collins, Colorado, USA \\
pouchet@colostate.edu}
}

\maketitle

\begin{abstract}
One of the challenges of using machine learning techniques with
medical data is the frequent dearth of source image data on which to
train. A representative example is automated lung cancer diagnosis,
where nodule images need to be classified as suspicious or benign. In
this work we propose an automatic synthetic lung nodule image
generator. Our 3D shape generator is designed to augment the variety
of 3D images. Our proposed system takes root in autoencoder
techniques, and we provide extensive experimental characterization
that demonstrates its ability to produce quality synthetic images.

\end{abstract}


\section{Introduction}
\label{sec:introduction}
Year after year, lung cancer is consistently one of the leading causes of cancer
deaths in the world \cite{Siegel17}. Computer aided diagnosis, where
a software tool proposes a diagnosis after analyzing the patient's medical imaging results,
is a promising direction: from an input
low-resolution 3D CT scan, image analysis techniques can be used to
classify nodules in the lung scan as benign or potentially cancerous.
But such systems require large amounts of labeled 3D training images to ensure
the classifiers are adequately trained with sufficient generality. 
Especially when new technologies are developed, cancerous lung nodule detection 
still suffers from a dearth of training images which hampers the ability to 
effectively improve and automate the analysis of CT scans for cancer risks
\cite{Valente16}. In this work, we propose to address this problem by automatically
generating synthetic 3D images of lung nodules, to augment the training
dataset of such systems with meaningful (yet computer-generated) images \cite{Wang19}.

Features of lung nodules computed from 3D images can be used as inputs
to a nodule classification algorithm.
Features such as volume, degree of compactness, surface area to volume ratio, etc. have
been useful in classifying lung nodules \cite{Li08}.
2D lung nodule images that are realistic enough to be classified by radiologists as
actual CT scan images have been created using generative
adversarial networks (GANs) \cite{Chuquicusma17}. In
our work, we aim to generate 3D lung nodule images which match the
feature statistics of actual nodules as determined by an analysis
program. We propose a new system inspired from autoencoders, and
extensively evaluate its generative capabilities. Precisely, we present in detail LuNG: a synthetic lung nodule generator, which is a neural network trained to generate new examples of 3D
shapes that fit within a broad learned category \cite{Wang19}.

To improve automatic classification in cases where input images are
difficult to acquire, our work aims to create realistic synthetic
images given a set of seed images. For example, the Adaptive Lung
Nodule Screening Benchmark (ALNSB) from the NSF Center for
Domain-Specific Computing \cite{cdsc-web} uses a flow that leverages
compressive sensing to reconstruct images from low-dose CT
scans. Compared to compressive sensing, filtered backprojection is a
technique which has more samples readily available (such as
LIDC/IDRI \cite{Rong17}), but filtered backprojection has slightly
different images that are not appropriate for training ALNSB.

For evaluation, we integrated our work with ALNSB
\cite{Shen15} that automatically processes a low-dose 3D CT
scan, reconstructs a higher-resolution image, isolates all nodules in
the 3D image, computes features on them and classifies each nodule as
benign or suspicious. We train LuNG using original patient data, and
use it to generate synthetic nodules ALNSB can process.
We create a network which optimizes 4 metrics: (1) increase the
percentage of generated images accepted by the nodule analyzer; (2)
increase the variation of the generated output images relative to the
limited seed images; (3) decrease the difference of the means for computed features
of the output images relative to seed images; and (4) decrease the autoencoder
reproduction error for the seed images.

We make the following contributions.
First, we present LuNG, a new automated system for the generation of synthetic 3D images that span the feature space of the input training images. It uses novel metrics for the numerical evaluation of 3D image generation aligned with qualitative goals related to lung nodule generation.
Second, we conduct an extensive evaluation of this system to generate 3D lung nodule images, and its use within an existing computer-aided diagnosis benchmark application including iterative training
techniques to further refine the quality of the image generator.

The rest of the paper is organized as
follows. Section~\ref{sec:motivation} briefly motivates our work and
design choices. Section~\ref{sec:contribution} describes the LuNG
system. Extensive experimental evaluation is presented in
Section~\ref{sec:results}. Related work is discussed in
Section~\ref{sec:related} before concluding.

\section{Motivation}
\label{sec:motivation}
To improve early detection and reduce lung cancer mortality
rates, the research community needs to improve lung nodule detection even given low
resolution images and a small number of sample images for training. The
images have low resolution because low radiation dosages allow for screening
to be performed more frequently to aid in early detection, but the low radiation
dosage limits the spatial resolution of the image. The number of training samples
is small due in part to patient privacy concerns but is also related to the rate
at which new medical technology is being created which generates a need for new
training data on the new technology. Our primary goal is to create 3D voxel images
that are within the broad class of legal nodule shapes that may be generated from a
CT scan.

With the goal of creating improved images for training, we evaluate
nodules generated from our trained network using the same software
that analyzes the CT scans for lung nodules. Given the availability of
'accepted' new nodules, we test augmenting the training set with these
nodules to improve generality of the network. The feedback process we
explore includes a nodule reconnection step (to insure final nodules
are fully connected in 3D space) followed by a pass through the
analyzer which will prune the generated set to keep 3D nodule feature
means close to the original limited training set. The need to avoid
overfitting the network for a small set of example images, as well as
learning a 3D image category by examples, guided many of the network
architecture decisions presented below.

While one goal of our work is to demonstrate the possibility to create a
family of images which have \emph{computed characteristics} (e.g.,
elongation, volume) that fit within a particular distribution range; another
goal is to generate novel images that are similar to an observed input nodule image.
Hence, in addition to creating a generator network, we shall 
create a feature network that can receive a seed image as input and
produce as outputs values for the generator that reproduce the seed image.
The goal of generating images related to a given input image motivates
our inclusion of the reconnection algorithm. Other generative networks
will prune illegal outputs as part of their use model
\cite{Cummins17}, but we wanted to provide more guarantee of valid
images when exploring the feature space near a given sample input.
The goal of finding latent feature values for existing images
leads naturally to an autoencoder architecture for our neural network.

Generative adversarial networks (GANs) \cite{li_sig17}
and variational autoencoders (VAEs) \cite{Doersch16}
are two sensible approaches to generate synthetic images from a
training set and could intuitively be applied to our problem.
However, traditional GANs do not provide a direct mapping from source
images into the generator input feature space \cite{GANGuide},
which limits the ability to generate images similar to a specific 
input sample or require possibly heavy filtering of
``invalid'' images produced by the network. In contrast, using an
autoencoder-based approach as we develop below allows to better
explore the feature space near the seed images.
A system that combines the training of an autoencoder with a
discriminator network such as GRASS \cite{li_sig17} would allow some
of the benefits of GAN to be explored relative to our goals. However,
our primary goal is not to create images that match the distribution of
a training set as determined by a loss function. As we
show in section~\ref{sec:metrics}, our goal can be summarized as
creating novel images that are within a category acceptable to an
automated nodule analyzer. As such, we strive to generate images that
are not identical to the source images but fit within a broad category
learned by the network.

A similar line of reasoning can be applied to VAEs relative to our
goals. Variational autoencoders map the distribution of the
input to a generator network to allow for exploration of
images within a distributional space. In our work, we tightly
constrain our latent feature space so that our training images map
into the space but the space itself may not match the seed
distribution exactly to aid in the production of novel images. Like
GANs, there are ways to incorporate VAEs into our framework, and to
some extent our proposed approach is a form of variational
autoencoder, although with clear differences in both the training and
evaluation loop, as developed below.
Our work demonstrates one sensible approach for a full end-to-end
system to create synthetic 3D images that can effectively cover the
feature space of 3D lung nodules reconstructed via compressive
sensing. 

\section{The LuNG System}
\label{sec:contribution}

The LuNG system is based on a neural network trained to produce
realistic 3D lung nodules from a small set of seed examples to help improve
automated cancer screening. To provide a broader range of legal images,
guided training is used in which each nodule
is modified to create 15 additional training samples. We call the initial
nodule set, of which we were provided 51 samples, the 'seed' nodules. The 'base'
nodules include 15 modified samples per seed nodule for a total of 816 samples.
The base nodules are used to train an autoencoder neural network with 3 latent feature
neurons in the bottleneck layer. The output of the autoencoder goes through a reconnection
algorithm to increase the likelihood that viable fully connected nodules are being
generated. A nodule analyzer program then extracts relevant
3D features from the nodules and prunes away nodules outside the range of interesting
feature values. We use the ALNSB \cite{Shen15} nodule analyzer and classifier
code for LuNG.
The accepted nodules are the final output of LuNG for use in classifier training
or medical evaluation. Given this set of generated images which have been
accepted by the analyzer, we explore adding them to the autoencoder training
set to improve the generality of the generator. We explore
having a single generator network or 2 networks that exchange training samples.

\subsection{Input images}
\label{sec:inputimg}

\begin{figure}[h!tb]
\centering
\includegraphics[width=0.45\textwidth]{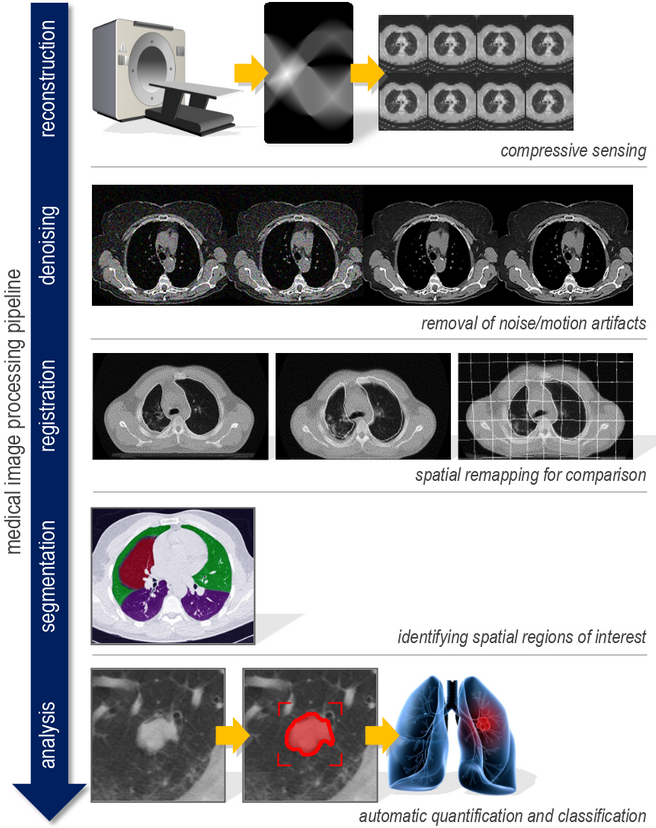}
\caption{Medical image processing pipeline \cite{cdsc-web}}
\label{fig:pipe}
\end{figure}

The input dataset comes from the Automatic Lung Screening Benchmark
(ALNSB) \cite{Shen15}, produced by the NSF Center for Domain-Specific
Computing \cite{cdsc-web}. This pipeline, shown in figure~\ref{fig:pipe}, is targeting
the automatic reconstruction of 3D CT scans obtained with reduced
(low-dose) radiation, i.e., reducing the number of samples taken by
the machine. Compressive sensing is used to reconstruct the initial 3D
image of a lung, including all artifacts such as airways, etc. A
series of image processing steps are performed to isolate all
nodules that could lie along the tissue. Then, each of the 3D
candidate nodules is analyzed to obtain a series of domain-specific
metrics, such as elongation, volume, surface area, etc. 
An initial filtering is done based on
static criteria (e.g., volume greater than $4 \text{mm}^3$) to prune
nodules that are not suspicious for potential cancerous
origin. Eventually, the remaining nodules are fed to a trained
SVM-based classifier, which classifies the nodules as potentially
cancerous or not. The end goal of this process is to trigger a
high-resolution scan for only the regions containing suspicious
nodules, while the patient is still on the table.

This processing pipeline is extremely specific regarding both the
method used to obtain and reconstruct the image, via compressive
sensing, and the filtering imposed by radiologists regarding the nodule
metrics and their values about potentially cancerous nodules. In this
paper, we operate with a single-patient data as input, that is, a
single 3D lung scan. About 2000 nodules are extracted from this single
scan, out of which only 51 are true candidates for the
classifier. Our objective is to create a family of nodules that are
also acceptable inputs to the classifier (i.e., which have not been
dismissed early on based on simple thresholds on the nodule metrics),
starting from these 51 images. We believe this 
represents a worst-case scenario where the lack of input images is not
even sufficient to adequately train a simple classifier and is
therefore a sensible scenario to demonstrate our approach to
generating images within a specific acceptable feature distribution.

These 51 seed images represent the general nodule shape that we
wish to generate new nodules from. Based on the largest of these 51
images, we set our input image size to 25$\times$28$\times$28mm, which
also aligns with other nodule studies \cite{Li08}. The voxel size
from the image processing pipeline is 1.25$\times$0.7$\times$0.7mm, so our
input nodules are 20$\times$40$\times$40 voxels. This results in an input to the
autoencoder with 32,000 voxel values which can range from 0 to 1.

\begin{figure}[h!tb]
\centering
\includegraphics[width=0.4\textwidth]{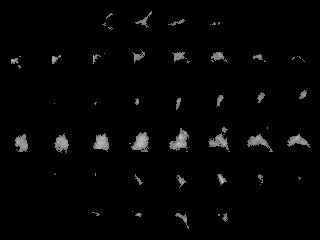}
\caption{Six of the 51 seed nodules showing the middle 8 out of 20 2D slices}
\label{fig:raw}
\end{figure}

Figure~\ref{fig:raw} shows 6 of the 51 seed images from the CT scan. 
Each of the images is centered in the 20$\times$40$\times$40
training size. One of our nodules was slightly too wide and 21 out of 1290 
total voxels were clipped; all other nodules fit within the training size. 
From an original
set of 51 images, 816 are generated: 8 copies of each nodule are the 8 possible
reflections in X,Y, and Z of the original; and 8 copies are the X,Y, and Z
reflections of the original shifted by 0.5 pixels in X and Y. The
reflections are still representative of legal nodule shapes to the analyzer,
so it improves the generality of the autoencoder
to have them included. The 0.5-pixel shift also aids generalization of the
network by training it to tolerate fuzzy edges and less precise pixel values. We do not
do any resizing of the images as we found through early testing that utilizing the full
voxel data resulted in better generated images than resizing the input and output of the autoencoder.

Our initial 51 seed images include 2 that are classified as suspicious nodules.
These 2 seed images become 32 images in our base training set, but
still provide us with a limited example of potentially cancerous nodules. A primary goal of the
LuNG system is to create a wider variety of images for use in classification based on learning
a nodule feature space from the full set of 51 input images.

\subsection{Autoencoder network}
\label{sec:autoencoder}
\begin{figure}[h!tb]
\centering
\includegraphics[width=0.45\textwidth]{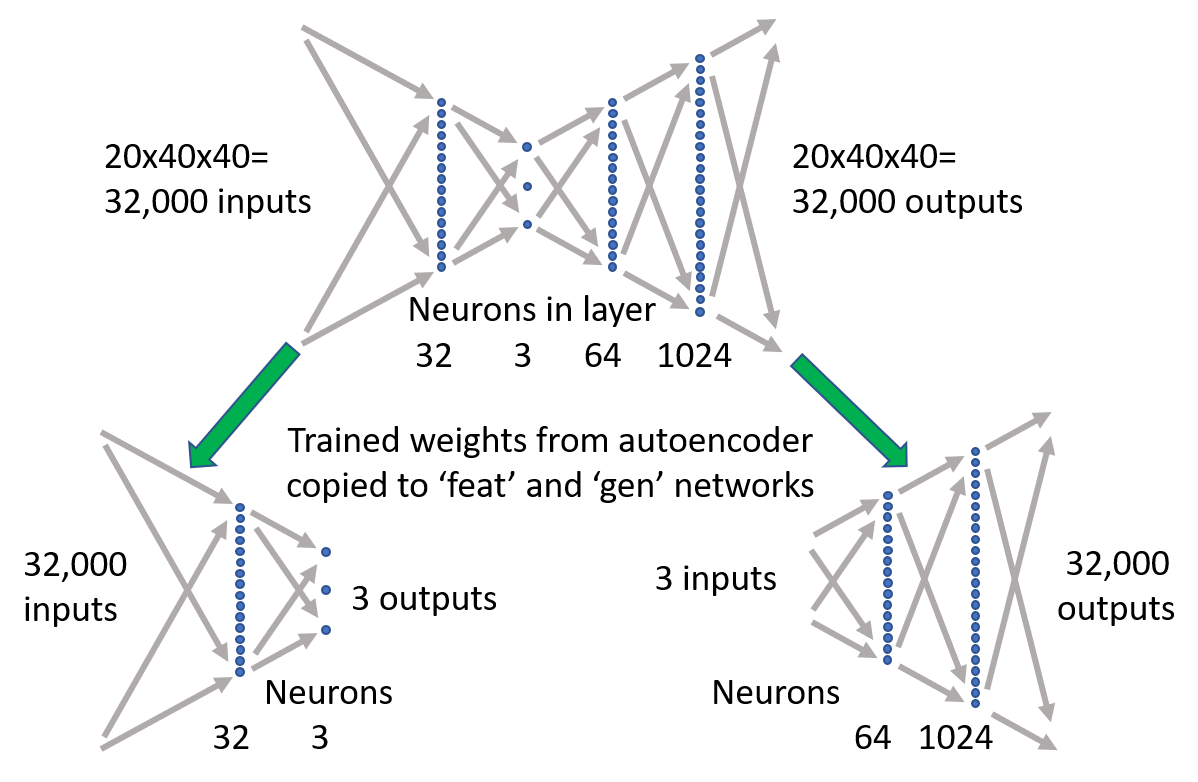}
\caption{Autoencoder and derived feature/generator networks for nodules}
\label{fig:autoencoder}
\end{figure}

Figure~\ref{fig:autoencoder} shows the autoencoder structure as well as
the feature and generator networks that are derived from it. All internal layers use tanh for
non-linearity, which results in a range of -1 to 1 for our latent feature
space. The final layer of the autoencoder uses a sigmoid function to keep
the output within the 0 to 1 range that we are targeting for voxel values.

We experimented with various sizes for our network and various methods
of providing image feedback from the analyzer with results shown in
section~\ref{sec:results}. The network
shown in figure~\ref{fig:autoencoder} had the best overall $Score$.

Our autoencoder is trained initially with the 816 images in our base set.
We use Adam \cite{Kingma14} for stochastic optimization to minimize the
mean squared error of the generated 32,000 voxel 3D images.
After creating a well-trained autoencoder, the network can be split into
feature and generator networks. The feature network can be used to map
actual nodules into a latent feature space so that novel images similar
to the actual input nodule can be created using the generator network.
If stepping is done between the latent feature values of nodule suspected
as cancerous and another
suspected to be non-cancerous, a skilled neurologist could identify the shape
at which the original suspicious nodule would not be considered suspicious
to help train and improve an automated classifier.
The generator network can also be used to generate fully random images for
improving a classifier. For our random generation experiments we use uniform values
from -1 to 1 as inputs for the 3 latent feature dimensions. We explore
the reasons and benefits of this random distribution in section~\ref{sec:features}.

The autoencoder structure which yielded the best results is not symmetric in that
there are fewer layers before the bottleneck layer than after. Like the seminal work by Hinton and Salakhutdinov \cite{Hinton06}, we explored various autoencoder sizes for our problem, but we added in an exploration of non-symmetric autoencoders. We found during hyperparameter testing
that a 2-layer feature network (encoder) performed better than a 1-layer or 3-layer network. 
We suspect that a single layer for the feature network was not optimal due to limiting the 
feature encoding of the input images to linear combinations of principle components \cite{Haykin09}. We suspect that 3 layers for our feature network was less optimal than 2 layers due to overfitting the model to our limited training set. Given our goal of generating novel nodule shapes, overfitting is a particular concern and we address this using a network scoring metric discussed in section~\ref{sec:metrics}.

\subsection{Reconnection algorithm}
\label{sec:reconnect}
The autoencoder was trained on single component nodules in that all
the 'on' voxels for the nodule were connected in a single 3D shape.
The variation produced by trained generator networks did not always result
in a single component, and it is common for generative networks that have
a technical constraint to discard output which fails to meet the requirements
\cite{Cummins17}. However, for the use case of exploring the feature space
near a known image, we chose to add a reconnection algorithm
to our output nodules to minimize illegal outputs. This algorithm insures that
for any input to the generative network, a fully-connected
nodule is generated.

When the generator network creates an image, a single fully-connected component is usually generated and the reconnection algorithm does not need to be invoked. 
In the case where multiple components are detected,
the algorithm will search through all empty voxels and set a small number of them to connect the components into a single nodule.

\subsection{Metrics for nodule analyzer acceptance and results scoring}
\label{sec:metrics}

The nodule analyzer and classifier computes twelve 3D feature values for
each nodule (features such as 3D volume, surface-to-volume ratio, and other
data useful for classification). Our statistical approach to this
data is related to Mahalanobis distances \cite{Mahalanobis36},
hence we compute the mean and standard deviations on these 12 features for the 51 seed nodules.
Random nodules from the generator are fed into
the classifier code and accepted to produce similar feature
values. This accepted set of images is the
most useful image set for further analysis or use in classifier training.

\paragraph*{Metrics for analyzer acceptance of the images}
Using the mean and standard deviation values we create a distance metric
$d$ based on concepts similar to the Mahalanobis distance.
Given $S$ is the set of 51 seed nodules and $i$ is the index for one of 12 features,
$\mu_{Si}$ is the mean value of feature $i$ and $\sigma_{Si}$ is the standard deviation.
Given Y is the set of output nodules from LuNG, the running mean for feature
$i$ of the nodules being analyzed is $\bar{Y_i}$. Given feature $i$ of a nodule $y$ is $y_i$
then if either (${y_i \geq \mu_{Si}}$ and $\bar{Y_i} \leq \mu_{Si}$) or ($y_i \leq \mu_{Si}$
and $\bar{Y_i} \geq \mu_{Si}$), then the nodule is accepted as it helps $\bar{Y_i}$ trend
towards $\mu_{Si}$. In cases where the nodule's $y_i$ moves $\bar{Y_i}$ away from $\mu_{Si}$,
we compute a weighted distance $d$ from $\mu_{Si}$ in multiples of ${\sigma_{Si}}$ using:

\[ d = |\frac{y_i + 3*\bar{Y_i} - 4*\mu_{Si}}{\sigma_{Si}}| \]

We compute the
probability of keeping a nodule $y$ as $P_{keep}$ which drops as $d$ increases:
{\small
\[ P_{keep} = \begin{cases}
  0.7+\frac{0.9}{d} & \text{if } y_i > \mu_{Si} \text{ and } \bar{Y_i} > \mu_{Si} \text { and } d > 3\\
  0.7+\frac{0.9}{d} & \text{if } y_i < \mu_{Si} \text{ and } \bar{Y_i} < \mu_{Si} \text { and } d > 3\\
  1  & \text{otherwise} \\

\end{cases} \]
}
The specific numerical values used for computing $d$ and $P_{keep}$ were chosen to
maximize the number of the original dataset which are accepted by this process
while limiting the deviation from the seed features allowed by the generator.
When using this process on a random sample from the 816 base nodules, 95\% were accepted.
Acceptance results for nodules generated by a trained network are provided in 
section~\ref{sec:results}.

\paragraph*{Metrics for scoring the accepted image set}
The composite score that we use to evaluate networks for LUNG is comprised of 4
metrics used to combine key goals for our work.
We compute the percentage of nodule images randomly generated by the generator
that are accepted by the analyzer.
For assessing the variation of output images relative to the seed images, we compute a 
feature distance ${FtDist}$ based on the 12 3D image features used in the analyzer.
To track how well the distribution of output images matches
the seed image variation, we compute a ${FtM\!M\!S\!E}$ based on the image feature
means. The ability of the network to reproduce a given seed image
is tracked with the mean squared error of the image output voxels,
as is typical for autoencoder image training.

Our metric of variation, ${FtDist}$, is the average distance over all accepted images
to the closest seed image in the 12-dimensional analyzer feature space and is scaled in
a way similar to Mahalanobis distances.
As ${FtDist}$ increases, the network is generating images that are less
similar to specific samples in the seed images, hence it is a metric we
want to increase with LuNG.
Given an accepted set of $n$ images $Y$ and a set of 51 seed images $S$,
and given $y_i$ denotes the value of feature $i$ for an image and $\sigma_{Si}$
denotes the standard deviation of feature $i$ within $S$:

\[ FtDist = 1/n\sum_{y \in Y}\min_{s \in S}\sqrt{\sum_{i=1}^{12}(\frac{y_i - s_i}{\sigma_{Si}})^2} \]

$FtM\!M\!S\!E$ tracks how closely LuNG is generating images that are within the same
analyzer feature distribution as the seed images. It is the difference between the means 
of the images in $Y$ and $S$ for the 12 3D features. As ${FtM\!M\!S\!E}$
increases, the network is generating images that are increasingly
outside the seed image distribution, hence we want smaller values for
LuNG. Given $\mu_{Si}$ is the mean
of feature $i$ in the set of seed images and $\mu_{Yi}$ is the mean
of feature $i$ in the final set of accepted images:

\[ FtM\!M\!S\!E = 1/12\sum_{i=1}^{12}(\frac{\mu_{Yi} - \mu_{Si}}{\sigma_{Si}})^2 \]

$Score$ is our composite network scoring metric used to compare different
networks, hyperparameters, feedback options, and reconnection options.
In addition to $FtDist$ and $FtM\!M\!S\!E$, we use $AC$, which is
the fraction of generated images which the analyzer accepted, and
$M\!S\!E$ which is the mean squared error that results when
the autoencoder is used to regenerate the 51 seed nodule images.

\[ Score = \frac{FtDist-1}{(FtM\!M\!S\!E+0.1)*(M\!S\!E+0.1)*(1-AC)} \]

$Score$ increases with ${FtDist}$ and ${AC}$ 
and decreases with ${FtM\!M\!S\!E}$ and $M\!S\!E$. The constants in
the equation are based on qualitative assessments of network results;
for example, using ${M\!S\!E+0.1}$ means that $M\!S\!E$ values below 0.1
don't override the contribution of other components and
aligns with the qualitative statement that an
$M\!S\!E$ of 0.1 yielded visually acceptable images in comparison with
the seed images.

Results using $Score$ to evaluate networks and LuNG interface features are 
discussed further is section~\ref{sec:results}. 
Our use of $Score$ to evaluate the entire nodule generation process rates the
quality of the random input distribution, the generator network, the reconnection
algorithm, the analyzer acceptance, and the interaction of these components
into a system. Our use of the analyzer acceptance rate is similar in some
functional respects to the discriminator network in a GAN as both techniques
are used to identify network outputs as being inside or outside an acceptable distribution.

\subsection{Updating the training set}
\label{sec:training}

\begin{figure}[h!tb]
\centering
\includegraphics[width=0.45\textwidth]{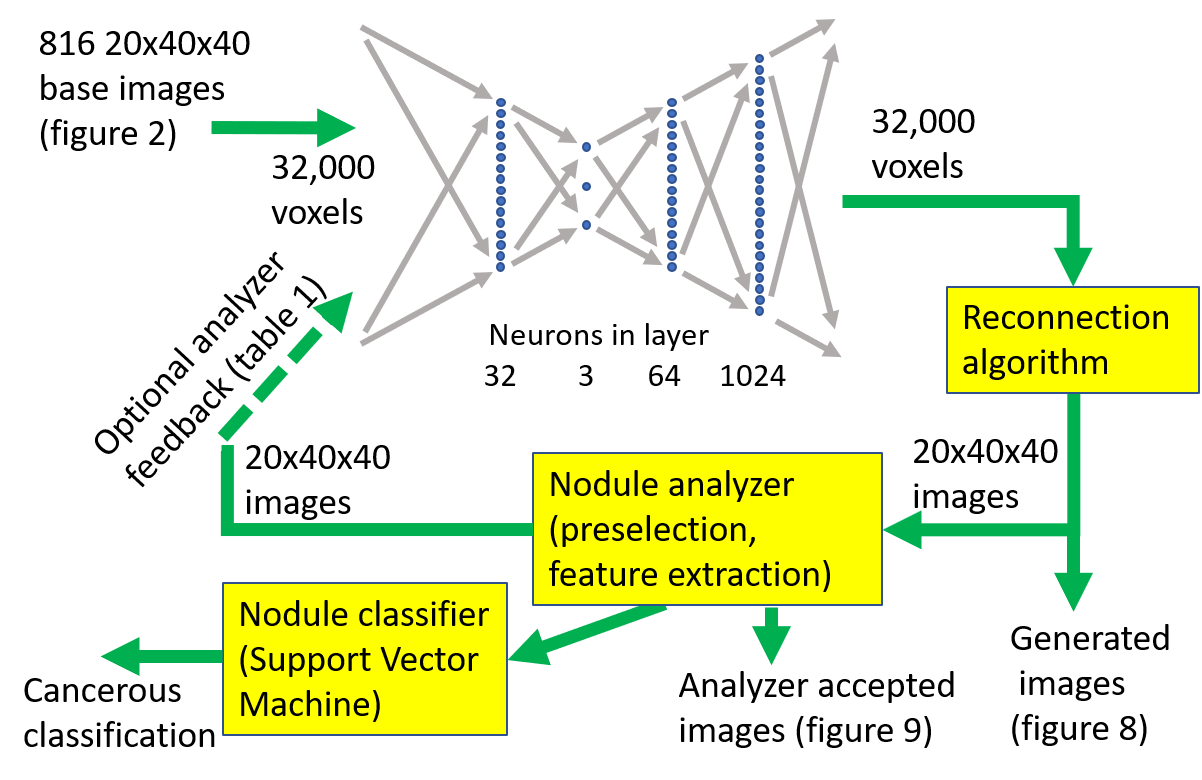}
\caption{Interaction between trained autoencoder and nodule analyzer. The images from figure~\ref{fig:raw} are always part of the training set to the autoencoder. The reconnected images after the network can be seen in figure~\ref{fig:step}. The analyzer accepted output of LuNG can be seen in figure~\ref{fig:final}.}
\label{fig:analyzer}
\end{figure}

After a trained generator network produces images which are reconnected and
validated by the nodule analyzer, a new training set may optionally be created
for the next round of training iterations of the autoencoder. 
Figure~\ref{fig:analyzer} diagrams
the data loop between the autoencoder and the nodule analyzer. We explored
various approaches for augmenting the training set, but ultimately
found that our best results came from proper autoencoder sizing and training
with only the 816 base images created by adding guided training examples to the original
51 seed images. Although the image feedback into the training set did not improve
the LuNG system, we include the description and results to
illustrate the drawbacks of the approach.

We were motivated to explore augmentation approaches because we wanted to
learn if adding accepted nodules to the training set could
improve network $Score$ results by improving $FtDist$. One of the approaches used images from
one trained network to augment a second trained network, similar
to the multi-adversarial networks discussed in \cite{Durugkar16}.
The intent of this approach is to improve the breadth of the feature space
used to represent all legal nodules by importing images that
were not generated by the network being trained.

We analyze the feedback approach we considered the best in
detail in section~\ref{sec:results}. In this approach, we train the network
for 50,000 iterations then generate 400 output images (less than half
the size of the base set). We pass those
images through the analyzer to insure they are considered legal and
chose a single random reflection of each image to add to the training set.
Because we are feeding in the image reflection, we did
not find added value in having 2 networks trade generated images - the
network itself was generating images whose reflections were novel to
both its own feature network as input and its generator network as output.
This training feedback is only done for 2 rounds of 25,000 training
iterations and then a final 50,000 training iterations train only on the
816 base images.

The augmentation experiments explored whether having
some training image variation helps fill the latent feature
space with more variation on legal images.
The intent of testing multiple approaches was to learn if an analyzer feedback
behavior can be found that improves the criteria LuNG is trying to
achieve: novel images accepted by the analyzer.

\section{Experimental Results}
\label{sec:results}

Using the $Score$ metric and its components introduced in section
~\ref{sec:metrics}, we evaluated various network sizes, the analyzer feedback
approaches discussed in section~\ref{sec:training}, and the reconnection
algorithm discussed in section~\ref{sec:reconnect}. Our naming for networks
is given with underscores separating neuron counts between the fully connected
layers, so 32\_3\_64\_1024 is a network with 4 hidden layers that have
32, 3, 64, and 1024 neurons respectively. As seen by
referencing tables~\ref{tab:params} and~\ref{tab:params2},
depending on the metric which is rated as most important, different
architectures would be recommended.

\subsection{Results for reconnection, feedback options, and depth}
\label{sec:optresults}

Table~\ref{tab:params} shows training results for some of the feedback
options discussed in section~\ref{sec:training}. The networks in this
table did not use the reconnection algorithm discussed in 
section~\ref{sec:reconnect}, so 'illegal' inputs to the analyzer could 
occur, reducing the acceptance rate. The MSE column shows 1000
times the mean squared error per voxel for the autoencoder when given the 51
original seed nodules as inputs and targets. The "AC\%" column shows
what percentage of 400 images randomly generated by the
generator network were accepted by the analyzer. The "FtDist" column shows the
average minimum distance in analyzer feature space from any image to a seed
image. The "FtMMSE" column shows the average mean squared error of all 12
analyzer features between the images and the 51 seed images. "No reflections"
is one of our feedback options referring to using the accepted images from 
the analyzer directly in the training set. "multiple" feedback refers to 
using 2 autoencoders and having each autoencoder use the accepted images 
that were output by the other. Using this table and other early results,
we observed that the network with no analyzer feedback had overall good
metrics, although the FtDist column indicating the novelty of
images generated was lower than we would prefer, so we weighed ${FtDist}$
heavier in our final scoring of networks as we explored network sizing.

\begin{table}
\caption{Key metrics for networks without reconnection}
{\begin{tabular}{ |p{3.5cm}||p{0.6cm}|p{0.6cm}|p{0.7cm}|p{1.0cm}| }
 \hline
 \multicolumn{5}{|c|}{Network parameter testing (2 run average after 6 rounds)} \\
 \hline
 Parameters & AC\% & MSE & FtDist & FtMMSE \\
 \hline
 16\_4\_64\_256\_1024 & 54 & 0.03 & 1.75 & 0.07 \\
 No Feedback      &  &  &  & \\
 \hline
16\_4\_64\_256\_1024  & 44 & 0.03 & 2.23 & 0.21 \\
FB: no reflections    &  &  &  & \\
\hline
16\_4\_64\_256\_1024  & 36  & 0.05  & 2.43 & 0.26 \\
FB: no reflections, multiple  &  &  &  & \\
\hline
16\_4\_64\_256\_1024  & 29  & 0.04  & 2.68 & 0.45 \\
FB: 4 reflections, multiple  &  &  &  & \\
\hline
\end{tabular}}
\label{tab:params}
\end{table}

Table~\ref{tab:params2} shows experiments in which the reconnection
algorithm was used. When using the reconnection algorithm,
the analyzer always has a full set of 400 images to consider for
acceptance, leading to higher acceptance rates.
This table includes data on the number of raw generator output images
which were clean when generated (one fully connected component) and the
number that were inverted (white background with black nodule shape).
The fact that deeper generation networks sometimes resulted in inverted
output images is an indication that they have too many degrees of freedom
and contributed to the decision to limit the depth of our autoencoder.
The "1 reflection" feedback label refers to having a single reflected
copy of each accepted image used to train the autoencoder for 2 of
the 6 rounds. This "1 reflection" feedback was our most promising
approach as described in section~\ref{sec:training}.

\begin{table}
\caption{Key metrics for networks using reconnection algorithm}
{\begin{tabular}{ |p{3.9cm}||p{0.45cm}|p{0.4cm}|p{0.55cm}|p{0.5cm}|p{0.5cm}| }
 \hline
 \multicolumn{6}{|c|}{Network parameter testing (2 run average after 6 rounds)} \\
 \hline
 Parameters & AC\% & MSE & FtDist & Clean & Invert \\
 \hline
64\_4\_64\_1024, No Feedback  & 85 & 0.08 & 1.78 & 109 & 0 \\
\hline
64\_4\_64\_1024, 1 reflection  & 64 & 0.06 & 3.13 & 63 & 0 \\
\hline
64\_4\_64\_256\_1024, No Feedback  & 80 & 0.02 & 1.96 & 117 & 2 \\
\hline
64\_4\_64\_256\_1024, 1 reflection  & 61 & 0.03 & 4.11 & 77 & 6.5 \\
\hline
\end{tabular}}
\label{tab:params2}
\end{table}

From the results in these 2 tables and other similar experiments,
we concluded that the approach in section~\ref{sec:training},
which used analyzer feedback for 2 of the 6 training rounds, had
the best general results of the 4 feedback approaches considered. Also,
the approach in section~\ref{sec:reconnect}, which will reconnect
and repair generator network outputs, yielded 3D images preferable
to the legal subset left when the algorithm was not applied. The
results of these explorations informed the final constants that
we used to create the $Score$ metric for rating networks as described
in section~\ref{sec:metrics}.

\subsection{Results for tuning network sizes}
\label{sec:sizresults}

\begin{figure}[h!tb]
\centering
\includegraphics[width=0.45\textwidth]{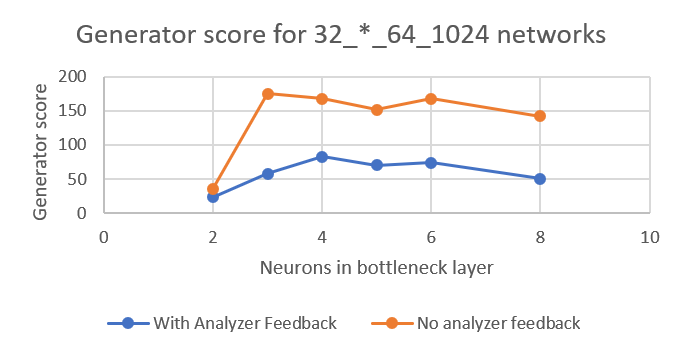}
\caption{$Score$ comparisons between networks that use 816 base images with
no analyzer feedback for 150,000 training iterations versus networks that trained for
25,000 iterations on the base images, then added 302 generated nodules to train for
25,000 iterations, then added a different 199 generated nodules to train for 25,000
iterations, and then finished with 75,000 training iterations with no feedback.}
\label{fig:scorefb}
\end{figure}

We analyzed neuron counts as well as total number of network layers using
$Score$. We tested multiple values for the neuron counts in each layer and
figure~\ref{fig:scorefb} shows the results for testing the dimensions in 
the latent feature space.
As can be seen, from the networks tested, the network which yielded
the highest score of 176 was 32\_3\_64\_1024, which is the network used
to generate the nodule images shown in section~\ref{sec:imagerslt}.

Our final network can train on our data in an acceptable amount
of time. Even though our experiments gathered significant intermediate
data to allow for image feedback during training, the final
32\_3\_64\_1024 network can be trained in approximately 2 hours.
Our system for training has 6 Intel Xeon E5-1650 CPUs at 3.6GHz and
an Nvidia GeForce GTX 1060 6GB GPU. Those 2 hours break down as:
10 minutes for creation of 816 base images from 51 seed images,
80 minutes to train for 150,000 epochs on the images, 20
minutes to generate and connect 400 nodules, and 10 minutes
to run the analyzer on the nodules. Code tuning would be able
to improve the image processing parts of that time, but the
training was done using PyTorch \cite{Pytorch} on the GPU and is already
highly optimized. When generating images for practical use,
we would recommend training multiple networks and using the results from
the network that achieved the highest score.

\begin{figure}[h!tb]
\centering
\includegraphics[width=0.45\textwidth]{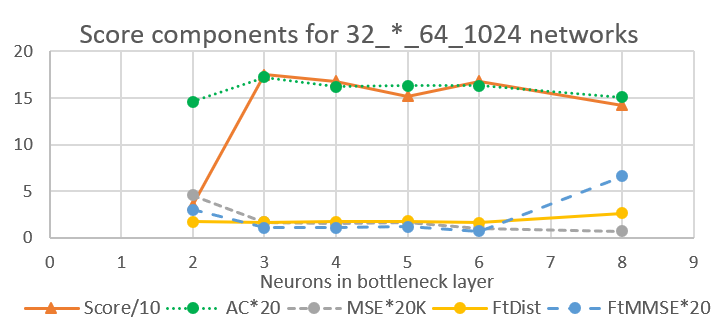}
\caption{There are 4 components used to compute the network score. The
component values are scaled as shown so that they can all be plotted on the same scale. }
\label{fig:metrics}
\end{figure}

Figure~\ref{fig:metrics} shows the components of $Score$ for the final
parameter analysis we did on the network. Note that
the MSE metric (mean squared error of the network on training set) continues
to decrease with larger networks, but $Score$ is optimal with
3 bottleneck latent feature neurons. Our intuition is that limiting
our network to 3 bottleneck
neurons results in most of the available degrees of freedom being required for
proper image encoding. As such, using a $-1$ to $1$ uniform random distribution as
the prior distribution for our generative network creates a variety of acceptable images. 
The $Score$ metric helps us to tune
the system such that we do not require VAE techniques to
constrain our random image generation process, although such techniques may be a valuable
path for future research. 

\subsection{Latent feature results}
\label{sec:features}

\begin{figure}[h!tb]
\centering
\includegraphics[width=0.5\textwidth]{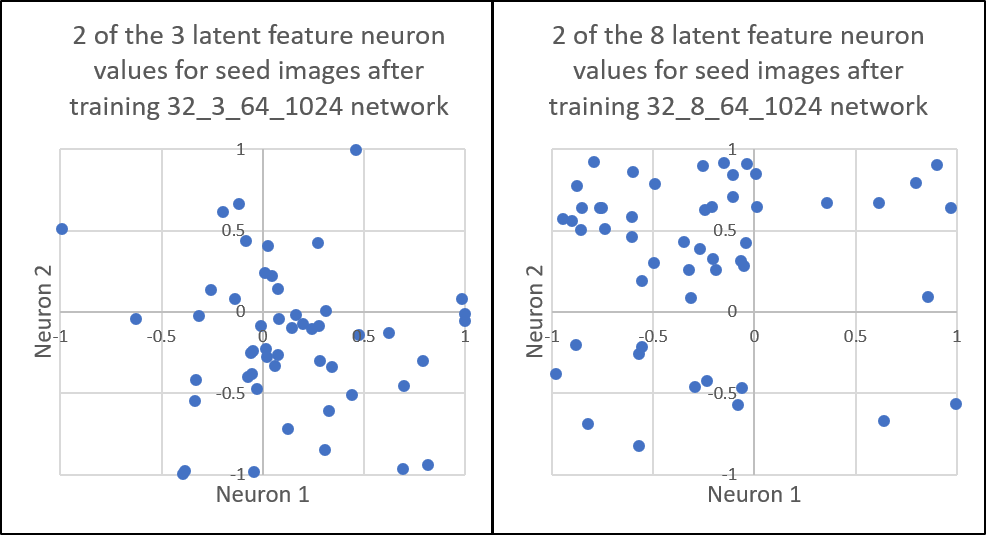}
\caption{Distribution of images in feature space after training with no feedback}
\label{fig:bottle}
\end{figure}

To visualize the weaknesses of too many bottleneck neurons in our
autoencoder network, we plot the 51 seed nodule positions in the latent
feature space in figure~\ref{fig:bottle}. To save space, we only 
present 2 of the 3 neurons for the final trained network, and we 
compare their positions to 2 of the 8 neurons from a trained network 
with 8 bottleneck neurons.

For the network with 3 bottleneck neurons, the plot shows that the
51 seed nodules are relatively well distributed in the 4 quadrants of
the plot and the full range of both neurons is used to represent all
the input images. For the network with 8 bottleneck neurons, most of
the seed nodules map to the upper left quadrant in the plot and the 
full range of the 2 neurons is not used. This is a symptom of having 
a network with more degrees of freedom than needed to represent the 
nodule training space. Our intuition is that this contributes to the 
high ${FtM\!M\!S\!E}$ measurements
shown in figure~\ref{fig:metrics} for a network with 8 bottleneck neurons.
The figures also show how a random -1 to 1 uniform range for our generator
will likely result in a higher acceptance rate for images generated with 
3 bottleneck neurons versus 8 bottleneck neurons.

\subsection{Image results}
\label{sec:imagerslt}
The images shown in this section are from a network with 4 hidden layers with
32, 3, 64, and 1024 neurons. The network was trained without analyzer feedback and
the output is processed to guarantee fully connected nodules.

\begin{figure}[h!tb]
\centering
\includegraphics[width=0.4\textwidth]{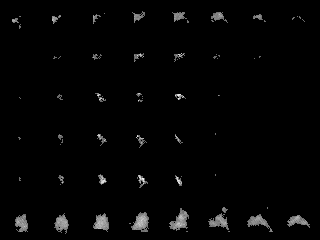}
\caption{6 steps through 3D latent feature space between original nodules 2 and 4 from figure~\ref{fig:raw}}
\label{fig:step}
\end{figure}

Figure~\ref{fig:step} shows the quality of 3D nodules our network can produce 
with 6 steps through the 3D bottleneck neuron
latent feature space starting at the 2nd nodule from figure~\ref{fig:raw} 
and ending at the 4th nodule. First, note that the
learned images for the start and end nodules are very similar to
the 2nd and 4th input images, validating the MSE data that the network
is correctly learning the seed nodules. The 4 internal step images
have some relation to the start and end images but depending on the
distance between the 2 nodules in latent feature space a variety of
shapes may be involved in the steps.

\begin{figure}[h!tb]
\centering
\includegraphics[width=0.4\textwidth]{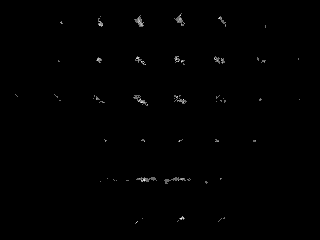}
\caption{6 images generated using uniform distribution from 3D feature space after passing nodule analyzer}
\label{fig:final}
\end{figure}

Figure~\ref{fig:final} shows 6 images generated by randomly selecting 3 values between
-1 and 1 for the latent feature inputs to the generator network and then being
processed by the analyzer to determine acceptance. When using the network
to randomly generate nodules (for classification by a trained
specialist or training automated classifiers), this is an example of quality final results.

\subsection{Image quality for training}
\label{sec:imagerqual}

\begin{figure}[h!tb]
\centering
\includegraphics[width=0.5\textwidth]{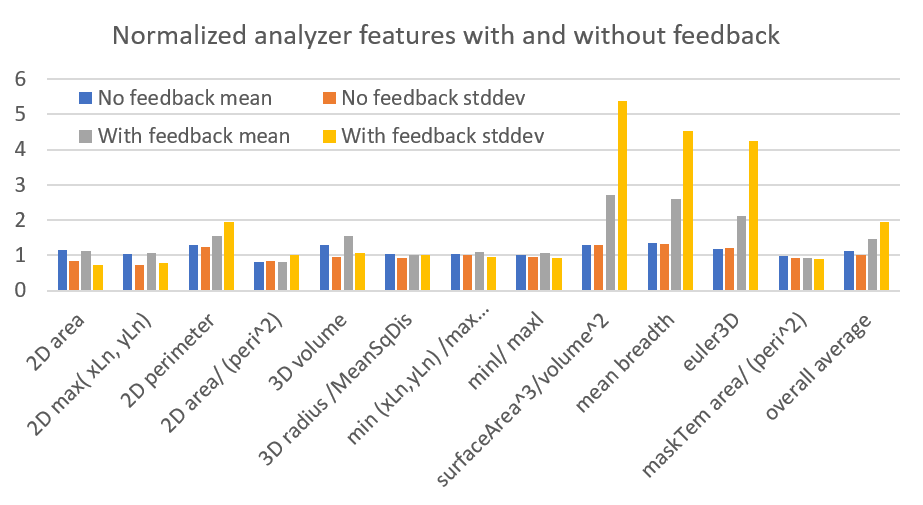}
\caption{Computed 3D features from nodule analyzer. Our proposed method avoids the deviations shown by the grey and yellow bars.}
\label{fig:all}
\end{figure}

Figure~\ref{fig:all} shows the 12 features that are used by the
nodule analyzer and demonstrates another key success of our full approach.
Characteristics like volume, surface area, and other values are used. 
We normalized the mean and standard deviation of each feature to 1.0 
and the figure shows that the mean and standard deviation of the 
generated nodules
for all 12 features stays relatively close to 1 for our proposed network with
no analyzer image feedback. However, when feedback is used, one can see that the
nodule features which have some deviation from the mean get amplified
even though the analyzer tries to accept nodules in a way that maintains
the same mean. For example, "surface area$^3$/volume$^2$" is a
measure of the compactness of a shape; the generated images from the network
with no feedback tended to have higher surface area to volume than
the seed images, and when these images were used for further training
the generated images had a mean that was about 2.6 times higher than
the seed images and a much higher standard deviation.

\begin{figure}[h!tb]
\centering
\includegraphics[width=0.45\textwidth]{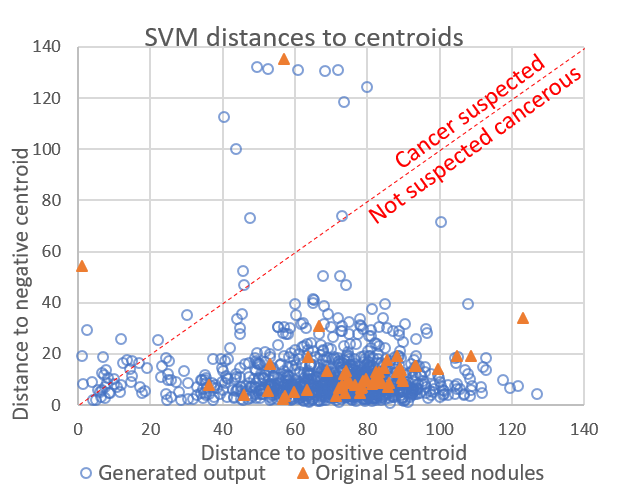}
\caption{Classifier distances to positive and negative centroids of SVM for 1000 analyzer accepted network generated samples. Nodules closer to positive than negative centroid after support vectors are applied are more likely cancerous.}
\label{fig:posneg}
\end{figure}

The ALNSB classifier we interact with uses a support vector machine to
map images onto a 2D space representing distances to positive (suspicious)
or negative (non-cancerous) centroids.
Figure~\ref{fig:posneg} shows the positive and negative centroid distances
for the seed data and 1000 samples of analyzer accepted generated data.
Nodules that have a smaller distance
to the positive centroid than to the negative centroid are classified
as likely cancerous. The general distribution of generated images fits
the shape of the seed images rather well, and there is a collection
of nodules being generated near the decision line between cancerous and
non-cancerous, allowing for improved training based on operator classification
of the nodules. Even though the original seed dataset only included 2 nodules 
eventually classified as potentially cancerous, our approach can use 
shape data from all 51 nodules to create novel images that can be useful for
improving an automated screening system. Note that these centroid distances
themselves are not part of the 12 features that are used to filter
the nodules, so this figure validates our approach to create
images usable for further automated classification work.

\section{Related Work}
\label{sec:related}
Improving automated CT lung nodule classification techniques and 3D
image generation are areas that are receiving significant research
attention.

Recently, Valente et al. provided a good overview of the requirements
for Computer Aided Detection systems in medical radiology and
they survey the status of recent approaches \cite{Valente16}. Our
aim is to provide a tool which can be used to improve the results of
such systems by both decreasing the false positive rate and 
increasing the true positive rate of 
classifiers through the use of an increase in nodules for training and
analysis. Their survey paper discusses in detail preprocessing,
segmentation, and nodule detection steps similar to those used in the
ALNSB nodule analyzer/classifier which we used in LuNG.

Li et. al provide a thorough overview of recent approaches to 3D
shape generation in their paper "GRASS: Generative Recursive
Autoencoders for Shape Structures" \cite{li_sig17}.
While we do not explore the design of an autoencoder with
convolutional and deconvolutional layers, the same image generation
quality metrics that we teach could be used to evaluate such designs. 
Tradeoffs between low reproduction error rates and overfitting would
have to be considered when setting the network depth and 
feature map counts in the convolutional layers.

Durugkar et al. describe the challenges of training GANs well and
discuss the advantages of multiple generative networks trained with
multiple adversaries to improve the quality of images generated
\cite{Durugkar16}. LuNG explored using multiple networks during
image feedback experiments. Larsen et al. \cite{Larsen15} teach 
a system which combines a GAN with an autoencoder which could be a basis for future
work introducing GAN methodologies into the LuNG system by
preserving our goal of generating shapes similar to existing
seed shapes.

\section{Conclusion}
\label{sec:conclusion}

To produce quality image classifiers, machine learning requires a
large set of training images. This poses a challenge for application
areas where large training sets are rare, such as for new medical
techniques using computer-aided diagnosis of cancerous lung nodules.

In this work we
developed LuNG, a lung nodule image generator, allowing us to augment the
training dataset of image classifiers with meaningful (yet
computer-generated) lung nodule images. Specifically, we have
developed an autoencoder-based system that learns to produce 3D images
with features that resemble the original training set.
LuNG was developed using PyTorch and is fully implemented and automated.
We have shown that the 3D nodules generated
by this process visually and numerically align well with the
general image space presented by the limited set of seed images.

\vspace{0.3cm}
\paragraph*{Acknowledgment}
This work was supported in part by the U.S. National Science Foundation award CCF-1750399.

\bibliographystyle{IEEEtran}
\bibliography{LuNG}

\end{document}